\documentclass[10pt,twocolumn,letterpaper]{article}

\usepackage{cvpr}
\usepackage{times}
\usepackage{epsfig}
\usepackage{graphicx}
\usepackage{amsmath}
\usepackage{amssymb}
\usepackage{xcolor,colortbl}

\definecolor{Blue1}{rgb}{0.95,0.95,0.95}
\definecolor{Blue2}{rgb}{0.98,0.98,0.98}

\usepackage[pagebackref=true,breaklinks=true,letterpaper=true,colorlinks,bookmarks=false]{hyperref}

 \cvprfinalcopy 

\def\cvprPaperID{****} 

\ifcvprfinal\fi
\begin{document}

\title{Face Alignment with Cascaded Semi-Parametric Deep Greedy Neural Forests}

\author{Arnaud Dapogny$^1$\\
{\tt\small arnaud.dapogny@isir.upmc.fr}
\and
Kevin Bailly$^1$\\
{\tt\small kevin.bailly@isir.upmc.fr}
\and
S\'{e}verine Dubuisson$^1$\\
{\tt\small severine.dubuisson@isir.upmc.fr}\\\\
$^1$ Sorbonne Universit\'{e}s, UPMC Univ Paris 06, CNRS, ISIR UMR 7222, 4 place Jussieu 75005 Paris\\
}

\maketitle

\begin{abstract}

Face alignment is an active topic in computer vision, consisting in aligning a shape model on the face. To this end, most modern approaches refine the shape in a cascaded manner, starting from an initial guess. Those shape updates can either be applied in the feature point space (\textit{i.e.} explicit updates) or in a low-dimensional, parametric space. In this paper, we propose a semi-parametric cascade that first aligns a parametric shape, then captures more fine-grained deformations of an explicit shape. For the purpose of learning shape updates at each cascade stage, we introduce a deep greedy neural forest (GNF) model, which is an improved version of deep neural forest (NF). GNF appears as an ideal regressor for face alignment, as it combines differentiability, high expressivity and fast evaluation runtime. The proposed framework is very fast and achieves high accuracies on multiple challenging benchmarks, including small, medium and large pose experiments.
\end{abstract}

\section{Introduction}

\begin{figure}
\centering
\includegraphics[width=\linewidth]{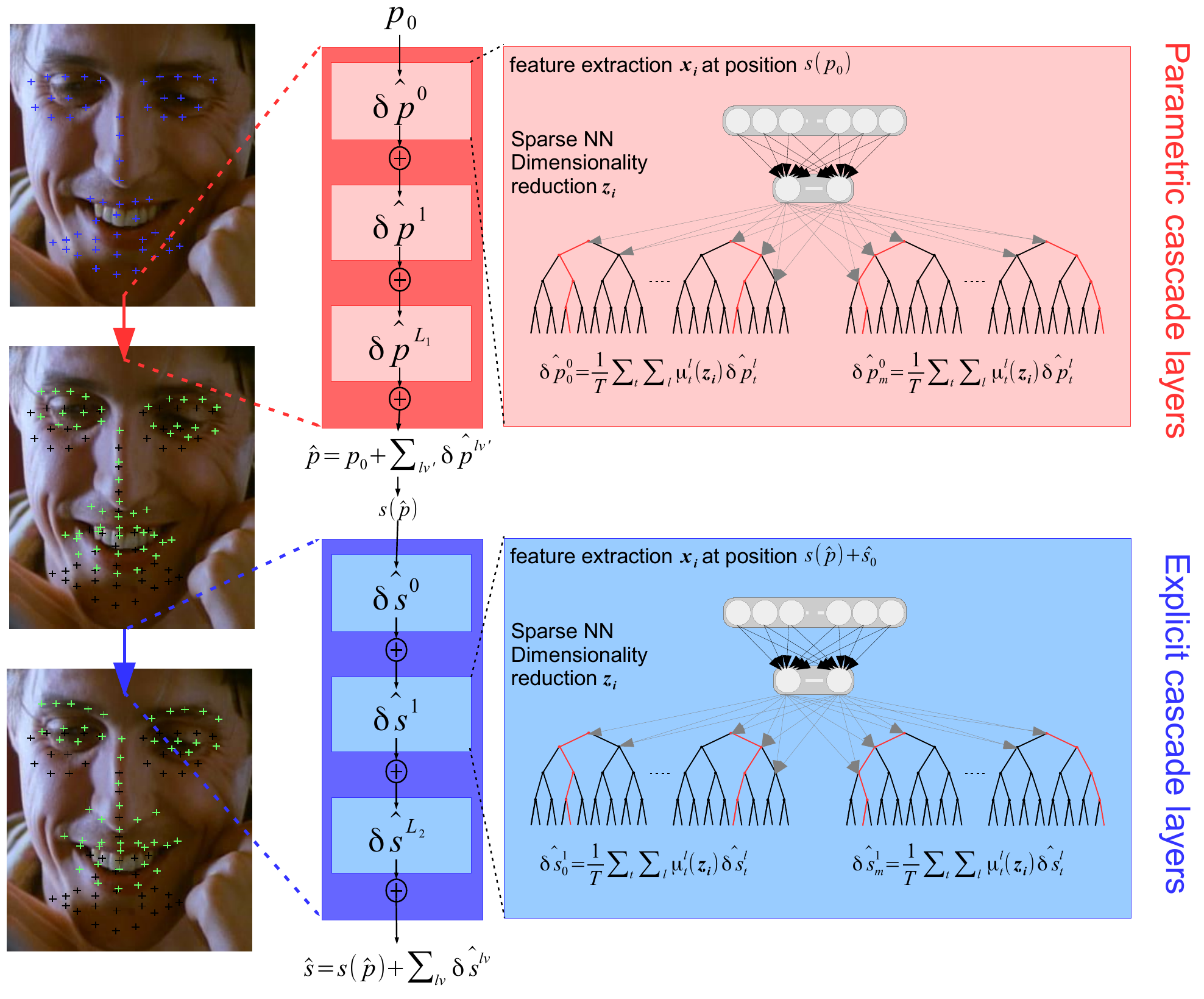}
\caption{Flowchart of the Cascaded Semi-Parametric deep GNF (CSP-dGNF) face alignment method. First, a parametric shape is regressed using deep GNFs. In the later stages, where the shape is closer to the target ground truth, more fine-grained deformations of the shape are estimated by means of an explicit regression.}
\label{flowchart2}
\end{figure}

Face alignment refers to the process of fitting a shape model on a face image, \textit{i.e.} precisely localizing keypoints that correspond, for instance, to eye or lip corners, nose tip or eyebrow locations. It serves as an input to many human-computer interaction systems, such as face reenactment \cite{thies2016face2face} or expression recognition \cite{dapogny2015pairwise}.

Most recent approaches for face alignment consist in applying a number of updates, starting from a mean shape. For each of these updates, a regressor is trained to map shape-indexed features to a displacement in either the space of a parametric face model (parametric regression), or directly in the space of the feature point locations (explicit regression). Then, given a face image, the shape is successively refined by applying the sequence of updates predicted by the learned regressors, in a cascaded way. On the one hand recent approaches such as \cite{asthana2014incremental}, \cite{tzimiropoulos2014gauss}, and \cite{tzimiropoulos2015project} advocate that the use of a parametric shape brings more stability to the alignment. Also, it allows to drastically reduce the dimensionality of the regression problem. On the other hand, approaches such as \cite{xiong2013supervised} and \cite{ren2014face} show that, given enough training data, explicit regression allows to capture fine-grained shape deformations, especially in the latter stages on the cascade. In our work, we propose to combine the best of those two worlds, by first applying updates in the space of a constrained, parametric model. Then, in the latter stages, fine-grained deformations can be captured by means of explicit regression layers (Figure \ref{flowchart2}).

Each individual cascade stage usually consist in (a) a dimensionality reduction step, and (b) a regression step. For instance, authors in \cite{xiong2013supervised}, \cite{tzimiropoulos2015project} and \cite{asthana2014incremental} use PCA to perform (a). In order to achieve (b), Xiong \textit{et al.} \cite{xiong2013supervised} use least-square error minimization. Asthana \textit{et al.} in \cite{asthana2014incremental} propose an incremental least-squares formulation. Martinez \textit{et al.} in \cite{martinez20162} use $\mathcal{L}_{2,1}$-norm regularization to induce robustness to poor initializations. These approaches offer the advantage of being very fast, especially when applying regression upon learned local features \cite{ren2014face}. However, performing (a) and (b) sequentially can lead to suboptimal solutions. Furthermore, those method generally use linear regressions to predict the updates. Hence the low number of parameters (which is constrained by the PCA output space dimension, in the case of SDM \cite{xiong2013supervised}) may hinder their ability to capture the variability of larger datasets, such as the one in \cite{zhu2015face}.

Moreover, deep learning techniques have recently started to show their capabilities for face alignment. In the work of Sun \textit{et al.} \cite{sun2013deep}, convolutional neural networks (CNNs) are used to extract suitable image representation. Then fully-connected layers are used to perform (a) and (b) in an end-to-end fashion. Zhang \textit{et al.} \cite{zhang2014coarse} use a cascade of deep autoencoders. Zhang \textit{et al.} \cite{zhang2016learning} design a deep learning pipeline to learn both (a) and (b) in a single pass. However, their approach require the use of large collections of images labelled with auxiliary attributes to help learn the image representations. Authors in \cite{zhang2016learning} also do not use cascaded regression, as the evaluation of several deep networks, and especially fully-connected layers, is prohibitively expensive in terms of computational load \cite{sun2013deep}, \cite{zhang2014coarse}.

Thus, one can wonder what could be an ideal candidate regressor in the frame of a cascaded regression pipeline to perform face alignment. Firstly, it shall be differentiable in order to learn (a) and (b) (and possibly the image representations) in an end-to-end fashion. Secondly, it shall be evaluated very fast in order to keep the runtime low when stacking several cascade layers. Thirdly, it has to embrace a good amount of parameters in order to scale well on larger databases, while at the same time it shall not overfit when trained on smaller corpses of examples.

Recently, Kontschieder \textit{et al.} \cite{kontschieder2015deep} introduced the deep neural forest (NF) framework for training differentiable decision trees. In this work, we propose to adapt the NF framework to achieve real-time processing. We call this method greedy neural forest (GNF) and wrap the deep GNF regressors inside a semi-parametric cascade. We demonstrate that the proposed cascaded semi-parametric deep GNF (CSP-dGNF) achieves high accuracies as well as very low runtime, while scaling very well to larger and more complex databases. To sum it up, the contributions of this paper are the following:

\begin{itemize}
\item a semi-parametric cascaded regression framework that combines the best of the two worlds, \textit{i.e.} a stable parametric alignment and a flexible explicit regression.
\item multiple improvements over NF, namely a greedy evaluation procedure to allow real-time processing, as well as a simplified training procedure involving constant prediction nodes. The proposed approach is flexible and tend not to overfit on training data.
\item a complete system that outperform most state-of-the-art approaches for face alignment while allowing very high framerates.
\end{itemize}

\section{Methodology}

\subsection{Semi-parametric cascade}

As it is somewhat classical in the landmark alignment literature, we propose a cascaded alignment procedure. However, in our work, we use a semi-parametric shape model, in which a shape prediction is provided as the sum of multiple displacements $\hat{\mathbf{\delta p}}^{(lv)}$ in parameter space, starting from an initial guess $\mathbf{p}_0$ (usually defined as the mean shape parameterization). Then, in the latter stages, the displacement is fine-tuned by applying explicit updates $\hat{\mathbf{\delta s}}^{(lv)}$. The final prediction can thus be written:

\begin{equation}\label{eq:fullprediction}
\begin{cases}
\ \mathbf{\hat{p}} = \mathbf{p}_0 + \sum_{lv'} \hat{\mathbf{\delta p}}^{(lv')} \\
\ \hat{\mathbf{s}} = s(\hat{\mathbf{p}}) + \sum_{lv} \hat{\mathbf{\delta s}}^{(lv)}
\end{cases}
\end{equation}

This allows a constrained shape regression that is, theoretically speaking, more stable than a fully explicit method, as well as a flexible procedure that captures the fine-grained feature point displacements (e.g. those who are related with facial expressions). After each step, the image descriptors are computed from the updated shape and used as inputs to the next cascade stage. 

\subsubsection{Parametric shape model}\label{paramshape}

Two constraints that may arise when using trees for the purpose of multi-output regression are (a) covering the output regression space by filling the leaf node predictions in a somewhat exhaustive manner and (b) limiting the number of nodes, which is a function of (the exponential of) the tree depth by the number of trees. Given those requirements, it is easy to see that trying to directly predict the shape displacement is a bad idea, as the output space is high-dimensional (dim. $N \times 2$ for a $N$-points markup) and the displacement value ranges can be important. For that matter, we use a shape parametrization in the first stages of the cascade, which is a classical setup for face alignment \cite{cootes1995active,cootes1998active,asthana2014incremental}. More specifically, shape $\mathbf{s}$ is defined as:

\begin{equation}\label{eq:parametericmodel}
\mathbf{\mathbf{s}(\mathbf{p})}=\alpha R(\gamma) (\mathbf{s}_0+\Phi \mathbf{g})+t
\end{equation}

Where $\alpha=(\alpha_x,\alpha_y)$ is a scaling parameter, $R$ is a $2D$ rotation matrix parametrized by angle $\gamma$, and $t=(t_x,t_y)$ is a translation parameter. Those are the rigid parameters of the transformation. $\mathbf{s}_0$ is the mean shape and vector $\mathbf{g}$ describes the non-rigid deformation of the shape in the space of the Point Distribution Model (PDM) $\Phi$, as it was introduced in the seminal work of Cootes \textit{et al} \cite{cootes1998active}. The vector of parameters is thus defined as $\mathbf{p}=(\alpha_x,\alpha_y, \gamma, t_x, t_y, g_1, ..., g_m) \in \mathbb{R}^{m+5}$. In our experiments, we set $m=15$, hence a 20-dimensional parametric shape.

For each training instance, we perform Procrustes analysis on the shape to remove the rigid component. Thus, we generate the PDM matrix $\Phi$ using PCA on the rigidly-aligned shapes. After that, we apply 100 Gauss-Newton iterations to retrieve the parameter vector $\mathbf{p}^*_i$ for image $i$ with ground truth shape $\mathbf{s}^*_i$. Each iteration is defined as $\mathbf{p}_i \leftarrow \mathbf{p}_i + (J(\mathbf{p}_i)^t J(\mathbf{p}_i))^{-1}J(\mathbf{p}_i)^t (\mathbf{s}^*_i-\mathbf{s}(\mathbf{p}_i))$, with $J(\mathbf{p}_i)$ the Jacobian of $s(\mathbf{p}_i)$.

\subsubsection{Explicit shape model.}\label{explicitshape}

Contrary to parametric layers (see Paragraph \ref{paramshape}), an explicit layer aims at directly predicting the displacements of the feature points. The output of such layer is defined as:

\begin{equation}\label{eq:explicitlayer}
\hat{\mathbf{\delta s}}=(\hat{\delta s_x^1}, ..., \hat{\delta s_x^{N}}, \hat{\delta s_y^1}, ..., \hat{\delta s_y^{N}}) \in \mathbb{R}^{N \times 2}
\end{equation}

As stated above, $\hat{\mathbf{\delta s}}$ is high-dimensional. Thus, for regressing these values using tree predictors, one shall restrict the ranges of the prediction values beforehand. In our case, the ranges of the deltas between a current predicted value and the ground truth feature point locations becomes smaller and smaller as cascade layers are stacked, allowing the use of explicit regression layers for the latter stages of the cascade.

\subsection{Regression with greedy Neural Forests}\label{MODGNF}

Within the frame of a cascaded landmark alignment \cite{xiong2013supervised}, it is crucial that each stage of the cascade (\textit{i.e.}, in our case, each NF predictor) does not overfit on the training data so that the residual deltas $\mathbf{\delta p}^*_i - \hat{\mathbf{\delta p}}_i$ (in the case a parametric layer) do not shrink too much after one or two stages. Even though the NF predictors embraces a whole lot of parameters ($20 \times 25 \times 255 \times 500 = 63750000$ parameters for a $20$-dimensional model and $T=25$ trees of depth $8$!), four mechanisms limit overfitting in practice:

\begin{itemize}
\item We use dimensionality reduction to limit the number of parameters (see Section \ref{reddim}).
\item We use early stopping by training each NF predictor with a restricted number of Stochastic Gradient Descent (SGD) updates. Moreover, as the proposed NF training framework is fully online, we generate on-the-fly random perturbations that are randomly sampled within the variation range for that parameter (for scaling and translation parameters only).
\item During training, optimization is performed only on the split nodes. The prediction nodes remain constant, enabling fully online training as well as reducing the computational load and the number of hyperparameters. Furthermore, suboptimal prediction node values can be compensated in the ulterior cascade layers.
\item When the training of a cascade layer is complete, we switch the NF predictor to its corresponding GNF.
\end{itemize}

As a result, the proposed approach appears to have very good properties w.r.t. overfitting. In fact, we demonstrate in Section \ref{eval} that the same cascade achieves low error rates both on small/medium poses data ($3148$ images for training) and large pose database ($61225$ images), with exactly the same hyperparameter setting and SGD optimization.

\subsubsection{GNF predictors}

\paragraph{Soft trees with probabilistic routing.}\label{softtrees} In the case of a classical decision tree, the probability $\mu^l(\mathbf{z_i})$ to reach leaf node $l$ given an example $\mathbf{z_i}$ is a binary function, that can be formulated as a product of Kronecker deltas that successively indicate if $\mathbf{z_i}$ is rooted left or right:

\begin{equation}\label{softprediction2}
\mu^l(\mathbf{z_i}) = \prod\limits_{n \in \mathcal{N}_l^{right}}{\delta^n(\mathbf{z_i})}\prod\limits_{n \in \mathcal{N}_l^{left}}(1-\delta^n(\mathbf{z_i}))
\end{equation}

Where $\mathcal{N}_l^{left}$ and $\mathcal{N}_l^{right}$ denotes the sets of nodes for which $l$ belongs to the left and right subtrees, respectively. Moreover, if we consider oblique splits:

\begin{equation}\label{deltatree}
\begin{cases}
        \delta^n(\mathbf{z_i}) &= 1 \quad\text{if}\quad \sum\limits_{j=1}^J{\beta^n_j z_{ij}} - \theta^n > 0\\
        \delta^n(\mathbf{z_i}) &= 0 \quad\text{if}\quad \sum\limits_{j=1}^J{\beta^n_j z_{ij}} - \theta^n \leqslant 0
 \end{cases}
\end{equation}

In the case of a Neural Forest (NF) \cite{kontschieder2015deep}, the probability $\mu^l$ to reach a leaf node $l$ is defined as a product of continuous split probabilities associated to each probabilistic split node $n$ (Equation \eqref{softprediction}), that are parametrised by Bernoulli random variables $d^n \in [0,1]$. Taking the expected value (which corresponds to an infinite number of samplings from tree $t$), an example $\mathbf{z}_i$ goes to the right subtree associated to node $n$ with a probability given by the activation function $d^n(\mathbf{z}_i)$, and to left subtree with probability $1-d^n(\mathbf{z}_i)$.

\begin{equation}\label{softprediction}
\mu^l(\mathbf{z_i}) = \prod\limits_{n \in \mathcal{N}_l^{right}}{d^n(\mathbf{z}_i)}\prod\limits_{n \in \mathcal{N}_l^{left}}(1-d^n(\mathbf{z}_i))
\end{equation}

The activation $d^n(\mathbf{z}_i)$ for node $n$ is defined as follows:

\begin{equation}\label{weightsdef}
d^n(\mathbf{z}_i) = \sigma(\sum\limits_{j=1}^k{\beta^n_j z_{ij}} - \theta^n)
\end{equation}

Thus, the calculus of $d^n(\mathbf{z}_i)$ can be seen as the activation of a neuron layer with weights $\{\beta^n_j\}$ and bias $-\theta^n$. From a decision tree perspective, the successive activations $d^n(\mathbf{z}_i)$ define a soft routing through the trees, where each leaf node $l$ is reached with probability $\mu^l$.

\paragraph{Online learning with recursive backpropagation.}

The prediction error $\epsilon^l_t$ for a ground truth value $\delta p^*$ (in the case of a parametric layer) and a leaf $l \in \mathcal{L}$ of tree $t$ can be computed as the Euclidean distance between this ground truth value and the leaf prediction $\hat{\delta p^l_t}$. The prediction error for the whole tree is thus equal to:

\begin{equation}\label{ndfepsilon}
\epsilon_t(\mathbf{z}_i)=\sum_l{\mu^l(\mathbf{z}_i)\epsilon^l_t}
\end{equation}

The same holds true for an explicit layer (with displacements in feature point space $\hat{\mathbf{\delta s}}$). Hence, for any parameter $\phi^n$ (\textit{i.e.} a feature weight $\beta^n_j$ or the threshold value $\theta^n$), the parameter update is given by Equation \eqref{parupdate} (with $\alpha$ the learning rate hyperparameter).

\begin{equation}\label{parupdate}
\phi^n \leftarrow \phi^n - \alpha \frac{\partial \epsilon_t(\mathbf{z}_i)}{\partial \phi^n}
\end{equation}

Moreover, the derivatives of $\epsilon_t$ w.r.t. the parameters of a split node $n$ can be calculated recursively. Specifically, for a split node $n$ we can split the sum in Equation \ref{ndfepsilon} in three, by grouping the leaves that belong to the left $\mathcal{L}(n)$ and right subtrees $\mathcal{R}(n)$, and those who do not belong to those subtrees:

\begin{multline}\label{partial1}
\epsilon_t(\mathbf{z}_i) =  \sum_{l \in \mathcal{L}(n)}{\mu^l(\mathbf{z}_i)\epsilon^l}
												+ \sum_{l \in \mathcal{R}(n)}{\mu^l(\mathbf{z}_i)\epsilon^l} \\
												+ \sum_{l \notin \mathcal{L}(n),l \notin \mathcal{R}(n)}{\mu^l(\mathbf{z}_i)\epsilon^l}
\end{multline}

While the first and second term respectively depend on $1-d^n(\mathbf{z}_i)$ and $d^n(\mathbf{z}_i)$, the last term does not depend at all on parameter $\phi^n$. We can thus write the partial derivatives of Equation \ref{partial1} as:

\begin{equation}\label{partial3}
\frac{\partial \epsilon_t(\mathbf{z}_i)}{\partial \phi^n} = \mu^n(\mathbf{z}_i)\frac{\partial d^n(\mathbf{z}_i)}{\partial \phi^n}(\epsilon^n_+(\mathbf{z}_i)-\epsilon^n_-(\mathbf{z}_i))
\end{equation}

With $\epsilon^n_-(\mathbf{z}_i)$ and $\epsilon^n_+(\mathbf{z}_i)$ the errors respectively for the left and right subtrees, and:

\begin{equation}\label{thetabetaupdates}
\begin{cases}
        \frac{\partial d^n(\mathbf{z}_i)}{\partial \theta^n} &= -d^n(\mathbf{z}_i)(1-d^n(\mathbf{z}_i)) \\
        \frac{\partial d^n(\mathbf{z}_i)}{\partial \beta_j^n} &= z_{ij} d^n(\mathbf{z}_i)(1-d^n(\mathbf{z}_i))
 \end{cases}
\end{equation}

Moreover, the error backpropagated up to node $n$ is:

\begin{equation}\label{NDFPbackprop}
\epsilon^n = d^n(\mathbf{z}_i)\epsilon^n_+(\mathbf{z}_i) + (1-d^n(\mathbf{z}_i))\epsilon^n_-(\mathbf{z}_i)
\end{equation}

and the error $\frac{\partial \epsilon_t(\mathbf{z}_i)}{\partial z_{ij}}$ corresponding to the $j^{th}$ component of an example $i$ that shall be backpropagated up to the feature level is:

\begin{multline}\label{dimredupdate}
       \frac{\partial \epsilon_t(\mathbf{z}_i)}{\partial z_{ij}} = \frac{1}{T \times (m+5)} \sum \limits_{t=1}^{T \times (m+5)} \sum \limits_{n} \\ \mu^n(\mathbf{z_i}) \beta_j d^n(\mathbf{z_i})(1-d^n(\mathbf{z_i}))(\epsilon^n_+(\mathbf{z_i})-\epsilon^n_-(\mathbf{z_i}))
\end{multline}

Once the trees are initialized, training samples are sequentially chosen from the data (SGD or mini-batch) and a forward pass through the trees provides the values of the probabilities $\mu^n(\mathbf{z}_i)$ and activations $d^n(\mathbf{z}_i)$ for each node $n$. For node $n$ the prediction error $\epsilon^n_-(\mathbf{z}_i)$ and $\epsilon^n_+(\mathbf{z}_i)$ respectively from the left and right subtrees. Parameters can thus be updated using Equations \ref{parupdate}, \ref{partial3} and \ref{thetabetaupdates}. Equation \ref{dimredupdate} provides an update to the error that can be backpropagated up to the feature level. Eventually, the updated error up to node $n$ can be obtained by applying Equation \ref{NDFPbackprop}.

The authors of \cite{kontschieder2015deep} suggest combining this split node optimization scheme to an update of the leaf probabilities, that they apply after a specific number of epochs using a convex optimization scheme while the split node parameters are left unchanged. However, in our case, we solely update the split nodes and prediction nodes remain constant during training. We found that performing optimization on split nodes only was sufficient to obtain satisfying accuracies on multiple classification and regression benchmarks. Furthermore, in the frame of a cascaded alignment, such setting effectively prevents overfitting as well as allowing a faster, fully online training procedure for each cascade stage. However, this requires careful initialization of the node predictions and tree depth hyperparameter.

\paragraph{Prediction node initialization.}

Indeed, the tree depth has to be chosen carefully in order to ensure a minimal ``resolution'' in term of leaf predictions. In the case of a parametric layer, we first have to estimate the mean $\bar{\delta}_k$ and standard variation $\sigma_k$ of the delta between the initial position in parameter space $\mathbf{\delta p}^0_k$ (that corresponds to the mean shape in the shape space, for the first level of the cascade) and the ground truth objective $\mathbf{\delta p}^*_k$, for each parameter $k$. We then generate $T$ single-objective trees for each parameter $k$ by assigning each leaf node $l$ a single prediction $\mathbf{\delta p}^l \sim \mathcal{N}(\bar{\delta}_k,\sigma_k)$. During training, all the $(m+5)$ model parameters are optimized jointly by updating each tree node with Equations \ref{parupdate}, \ref{partial3}, \ref{thetabetaupdates} using  a parameter-dependant learning rate $\alpha_k = \frac{\alpha_0}{\sigma_k}$ in order to take into account the discrepancies in the dynamics of the different model parameters. The same holds true for the explicit layers.

We provide in the Appendix of this paper a proof that, in the regression case with constant leaf predictions initialized from a gaussian distribution, we have a sufficient condition to have each value $y \in [\bar{\delta}_k - \sigma_k, \bar{\delta}_k + \sigma_k]$ close to at least one leaf node prediction $y_l$ (in the sense that $|y-y_l|<\epsilon$, with probability superior to $1 - \epsilon'$). We essentially show that this condition is satisfied if $\mathcal{D}>\mathcal{D}_0$, with

\begin{equation}\label{lowerbound}
\mathcal{D}_0 = \frac{1}{ln(2)}ln{\frac{ln(1-(1-\epsilon')^{\frac{1}{2\sigma_k}})}{ln(1-\frac{2\epsilon}{\sqrt{2\pi}\sigma_k}e^{-\frac{(\sigma_k+\epsilon)^2}{2\sigma_k^2}})}}
\end{equation}

In our case, setting $\mathcal{D}=8$ ensure that this condition is satisfied with $\epsilon'=0.01$ and $\epsilon_k=\sigma_k/10$ for all the ranges $\sigma_k$ (which experimentally vary from $10$ to $0.1$).

\paragraph{Greedy evaluation of NF}

If $d^n(\mathbf{x_i}) \rightarrow \delta^n(\mathbf{x_i})$ for each node and tree, the evaluation of a NF (Equation \ref{softprediction}) becomes similar to that of a decision forest with oblique splits (Equation \ref{softprediction2}). Intuitively, from a NF evaluation perspective, we successively choose the best path through the tree in a greedy fashion, node after node. Thus, we refer to this model as a Greedy Neural Forest (GNF).

On the one hand, in order to evaluate a NF that is composed of $T$ trees, we have to evaluate the probability to reach each leaf node of each tree. Consequently, each split node has to be evaluated, thus the complexity of applying NF to a $k$-dimensional input is $T.k.(2^{\mathcal{D}+1}-1)$, \textit{i.e.} exponential in the tree depth $\mathcal{D}$. By doing so, we essentially lose the runtime advantage of using ensemble of decision trees for prediction. In case of a GNF, on the other hand, only a single, locally ``best'' path through the $T$ trees has to be evaluated. Hence, its complexity is equal to $T.k.\mathcal{D}$, \textit{i.e.} linear in $\mathcal{D}$. Furthermore, as stated in \cite{kontschieder2015deep}, after training is complete, the split node activations of a NF are close to either $0$ or $1$, hence only very little noise is added by switching a NF to its corresponding GNF. However, switching the two is all the more relevant in the frame of a cascaded regression, as potential errors are compensated in further stages the cascade. Furthermore, it dramatically reduces the evaluation runtime, enabling real-time processing.

\subsubsection{Feature extraction and dimensionality reduction}\label{reddim}

We use SIFT features for their robustness and extraction speed. First, $9$-dimensional orientation bins and magnitude integral channels are computed for the whole face region of interest. Subsequently, SIFTs descriptors are generated for each feature point using its current position estimate and the integral channels, as it is explained in \cite{dollar2009integral}. For that matter, we use  $4 \times 4$ non-overlapping cells within a $40$-pixel window for each feature point. Finally, these descriptors are concatenated to form the initial shape descriptor $\mathbf{x_i}$.

Learning NFs with such high-dimensional descriptors would be quite slow in terms of memory and training time, let alone overfitting issues. For those reasons, as in \cite{xiong2013supervised} we perform dimensionality reduction. However, as stated above, as NF are differential classifiers, we can use a single, randomly initialized neuron layer to map the high-dimensionality descriptor $\mathbf{x_i}$ to a $J$-dimensional one $\mathbf{z_i}$. During training, we update the weights of that layers in a single, top-down, supervised training pass (as opposed to, e.g., applying PCA beforehand \cite{xiong2013supervised}). Optionally, we can use the truncated gradient algorithm introduced in \cite{langford2009sparse} to induce sparsity within the weights of the neuron layer (which will be refered to as the sparse NN), effectively reducing the computational load dedicated to dimensionality reduction. Using this setting, the update for a weight $w_{jj'}$ of the NN is computed as follows:

\begin{equation}\label{NNupdate}
w_{jj'} \leftarrow T(w_{jj'} - \alpha \frac{\partial \epsilon_t(\mathbf{z}_i)}{\partial z_{ij}}\frac{\partial z_{ij}(\mathbf{x}_i)}{\partial w_{jj'}} - \alpha \eta sgn(w_{jj'}), \Theta)
\end{equation}

Where $\alpha$ is the learning rate hyperparameter, $\eta$ denotes the relative importance of the $\mathcal{L}_1$ regularization, and $\frac{\partial \epsilon_t(\mathbf{z}_i)}{\partial z_{ij}}$ is the error residual of the prediction stage (Equation \ref{dimredupdate}). $T(v,\Theta)=0$ is the truncation operator, which outputs $0$ if $v<\Theta$, $v$ otherwise. As it will be shown in the experiments, using the sparse NN allows to drastically reduce the computational runtime while maintaining a good accuracy. 

\section{Evaluation}\label{eval}

\subsection{Experimental setup}\label{hyperparam}

Even though our method embraces a lot of hyperparameters, few of them are critical to the global accuracy of the system. The region of interest for each face image corresponding to the provided bounding box is cropped according to the bounding box position, resized to a $200 \times 200$ scale, and the mean shape is centered on that crop. Then a $4$-level cascade, with $3$ parametric layers ($T=25$ trees per parameter making $500$ trees total per layer) and $1$ explicit layer ($T=5$ trees per feature point coordinate), is applied for aligning the feature points. Tree depth is fixed to $8$, to ensure that the condition of Equation \ref{lowerbound} is satisfied for all parameters. The NN consists in $k=500$ output units with an hyperbolic tangent activation function, which seems a good trade-of between speed and accuracy. All weights for NN and NF are randomly initialized from a uniform distribution in the interval $[-0.01,0.01]$. These weights are optimized jointly with $200000$ SGD updates corresponding to randomly sampled images and perturbations in translation and scale, with a constant learning rate of $0.005$. For the sparse NN, we set $\eta=0.01$ and $\Theta=0.05$, which allows to zero out more than $90\%$ of the NN weights for each layer.

As for evaluation, we use the standard average point-to-point distance as the evaluation metric. As it is common in the literature, we report the mean accuracy normalized by the inter-pupil distance for $N=51$ and $N=68$ points mark-ups. For simplicity, we omit the '$\%$' symbol. For the large poses evaluation benchmark on $AFLW2000$-$3D$, we normalize the error using the bounding box size, as in \cite{zhu2015face}.

\subsection{Available data}

The \textbf{300W database} \cite{sagonas2013300} consists in $3$ datasets: AFW \cite{zhu2012face}, LFPW \cite{belhumeur2013localizing} and HELEN \cite{le2012interactive}. The HELEN database contains $2000$ images for training and $330$ images for testing. The LFPW database contains $811$ images for training and $224$ for testing. As it is done in the literature, we train our models on the concatenation of the training partitions of the LFPW and HELEN databases, as well as the whole AFW database, which makes a total of $3148$ training images. We evaluate our method on the test partition of LFPW, the test partition of HELEN (both constituting the common subset of 300W) as well as the challenging i-bug database ($135$ images) that contains several examples of partial occlusions as well as non-frontal head poses.

The \textbf{300W-LP database} is an extension of the 300W database that contains face images featuring extreme pose variations on the yaw axis, ranging from $-90$ to $+90$ deg. The database contains a total of $61225$ images obtained by generating additional views of the images from AFW, LFPW, HELEN and i-bug, using the algorithm from \cite{zhu2015face}.

The \textbf{AFLW2000-3D} dataset consists in fitted $3D$ faces and large-pose images for the first $2000$ images of the AFLW database \cite{kostinger2011annotated}. As it was done in \cite{zhu2015face}, we evaluate the capacities of our method to deal with non-frontal poses by training on 300W-LP and testing on AFLW2000-3D. This database consists of $1306$ examples in the $[0,30]$ yaw range, $462$ examples in the $[30,60]$ range and $232$ examples in the $[60,90]$ range. As in \cite{zhu2015face}, we report accuracy for each pose range separately, as well as the mean and standard deviation across those three pose ranges.

\subsection{Face alignment on small and medium poses}

\paragraph{Impact of semi-parametric alignment.}

\begin{figure*}
\centering
\includegraphics[trim=3cm 0 0 0,width=20cm]{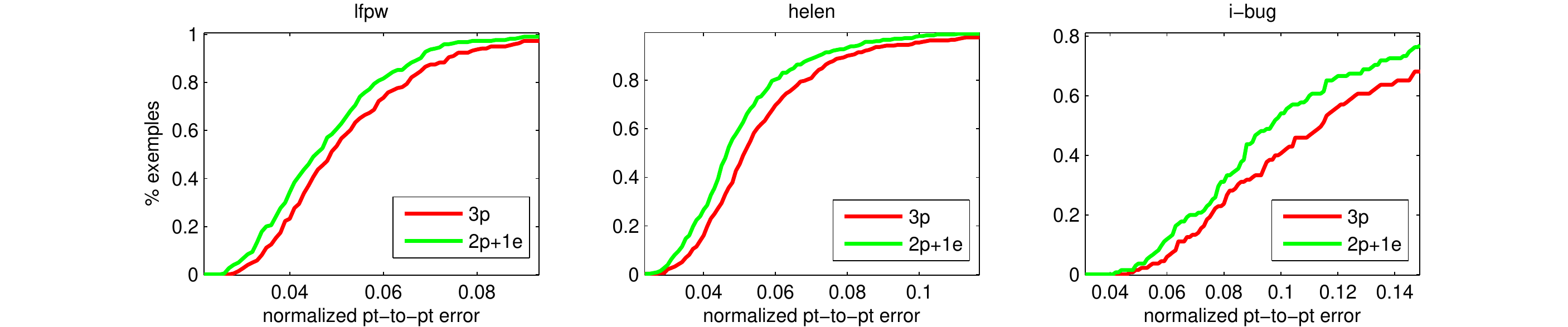}
\caption{Cumulative error distribution curves on the LFPW, HELEN and i-bug databases, for both a cascade with 3 parametric layers (3p) and 2 parametric layers + 1 explicit layer (2p+1e).}
\label{cds}
\end{figure*}

Figure \ref{cds} shows the cumulative error distribution curves on LFPW and HELEN test partitions, as well as on the i-bug database. More specifically, we study the impact of swapping the last layer of a $3$-layers parametric cascade with an explicit layer. As one can see, the error is generally lower for the semi-parametric cascade, notably for the most difficult examples on i-bug database. This shows that using an explicit layer allows to further decrease the error as compared to another parametric layer, as the last layer captures the fine-grained displacements between the ground truth and a well initialized parametric shape. Moreover, intuitively, using the parameters $\mathbf{p}^*_i$ estimated by Gauss-Newton optimization as a ground truth for alignment may induce some errors as compared to directly using  the ground truth shape $\mathbf{s}^*_i$ as a regression target. Hence, the addition of an explicit layer may help to circumvent this issue as well.

\paragraph{Comparison with state-of-the-art approaches.}

Table \ref{compalignment} shows a comparison of our approach with results reported for recent cascaded regression approaches. Noticeably, the accuracies reported for our $4$-levels CSP-dGNF are among the best results in the literature on the three databases, for both $N=51$ and $N=68$-landmark markups. This shows the interest of GNF as a predictor for cascaded regression systems, as well as the relevance of the semi-parametric approach.

\begin{figure*}[h!]
\centering
\caption{Examples of face alignment on small head poses from the HELEN database}
\label{smallposes}
\includegraphics[width=\linewidth]{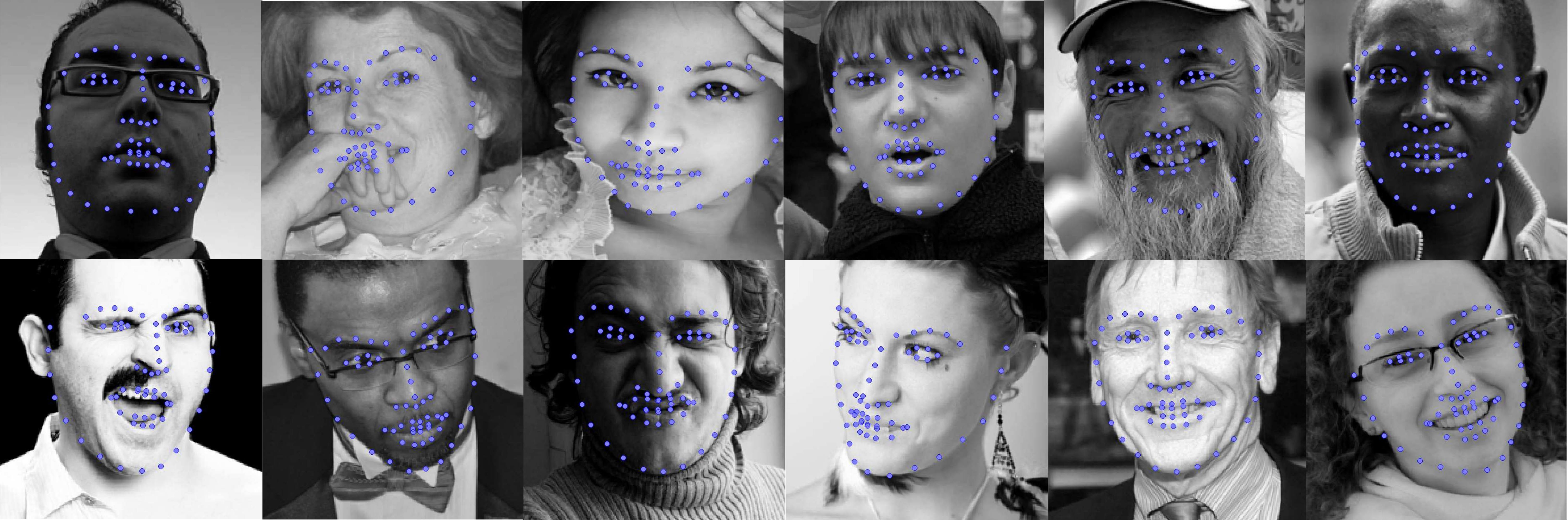}
\end{figure*}

\begin{figure*}[h!]
\centering
\caption{Examples of face alignment on medium head poses from the i-bug database}
\label{mediumposes}
\includegraphics[width=\linewidth]{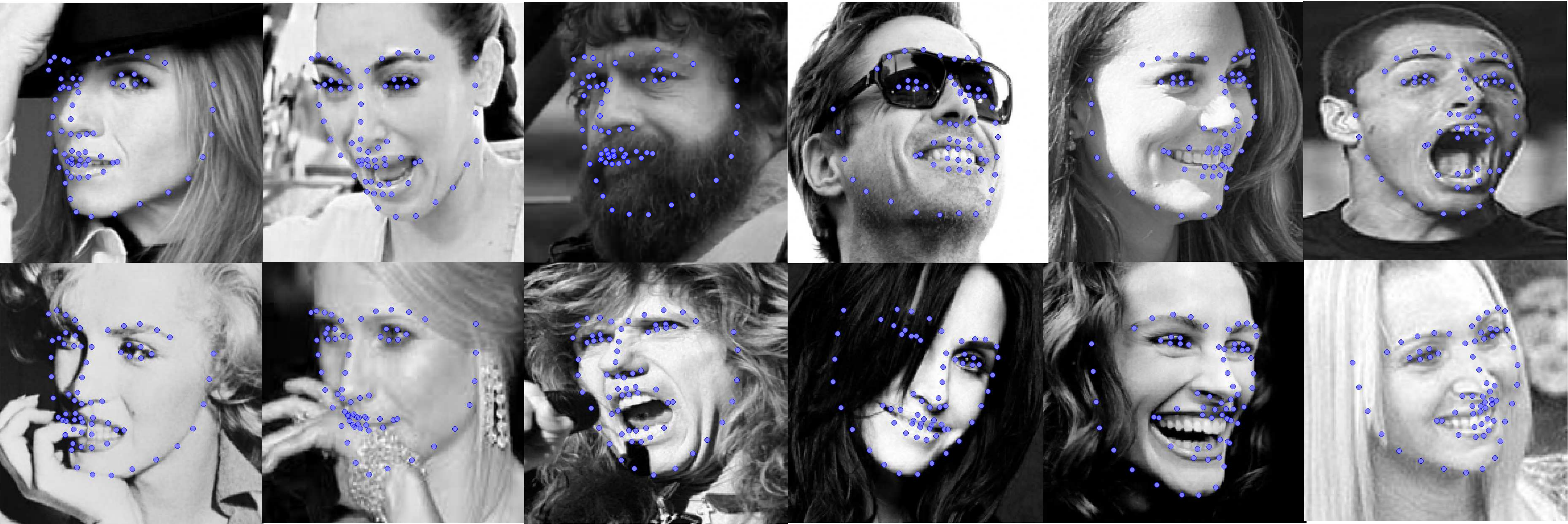}
\end{figure*}

Also note that our method performs similarly to the current best approach (TCDCN \cite{zhang2016learning}) on the common subset of 300-W, whereas TCDCN is more accurate on i-bug. However, authors of \cite{zhang2016learning} used $20000$ additional images labelled with auxiliary attributes for pretraining their network. Even though studying the impact of using deep representations is out of the scope of this paper, using pre-trained CNN layers as an input would be an interesting direction for future work. Indeed, GNF would allow to fine-tune upstream feature extraction layers in an end-to-end manner, similarly to how we train the NN. Figures \ref{smallposes} and \ref{mediumposes} displays examples of face alignment on small and medium poses, respectively.

\subsection{Face alignment on large poses}

Table \ref{compalignment3D} draws a comparison between our approach and recent face alignment methods on large pose data using the AFLW2000-3D database. Results obtained with other methods are gathered from \cite{zhu2015face}. First, in case where the training set is 300W (upper part of the table), the proposed CSP-dGNF achieves significantly higher accuracy than RCPR, ESR and SDM, for all three pose ranges.

Moreover, when trained on 300W-LP, CSP-dGNF also outperform those three methods by a wide margin. It is also more accurate than 3DDFA and 3DDFA+SDM for $[0,30]$, $[30,60]$ and $[60,90]$ yaw angles. Note that 3DDFA benefit from dense 3D alignment before regressing the 68-points $2D$ shape, and is much slower than ours: $75$ ms for 3DDFA using the GPU, plus SDM workload, whereas CSP-dGNDF largely runs in real time on a single CPU (See Section \ref{realtime}).

\begin{figure*}[h!]
\centering
\caption{Examples of face alignment on large  poses from the AFLW2000-3D database}
\label{largeposes}
\includegraphics[width=\linewidth]{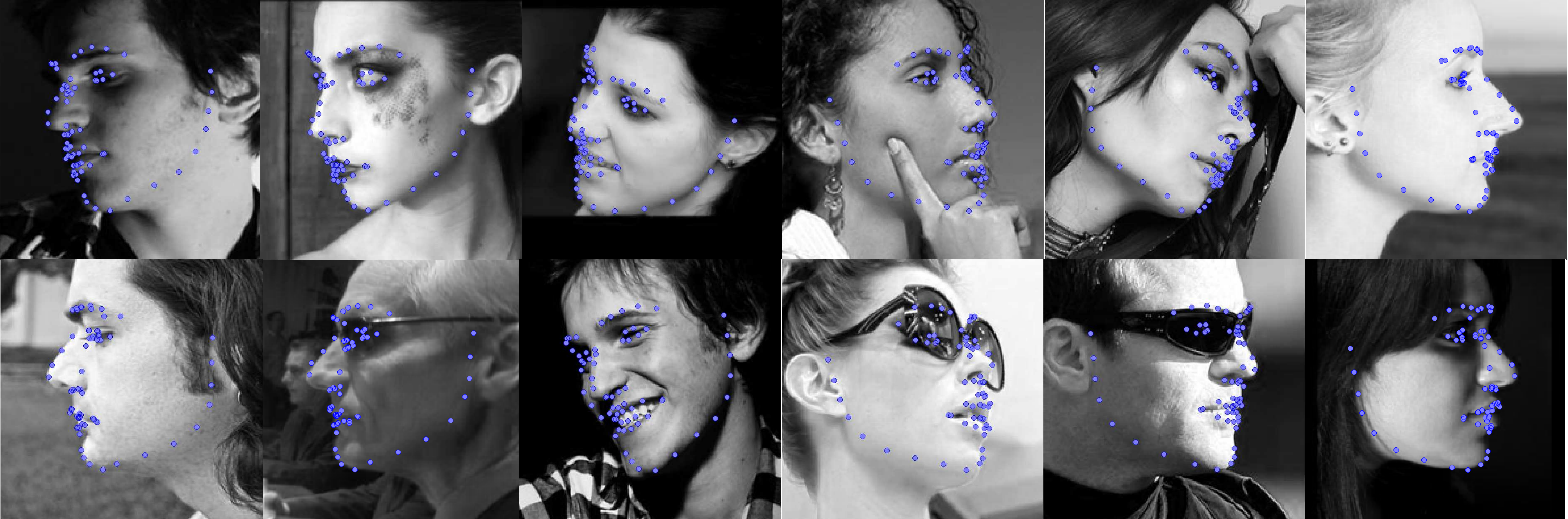}
\end{figure*}

Interestingly, the accuracies reported on Table \ref{compalignment3D} where obtained using the exact same hyperparameters that were used for alignment on small and medium poses. Our approach is also fully two-dimensional, and therefore would greatly benefit from using a $3D$ parametric model for large pose alignment. Hence, we believe there is considerable room for improvement. However, as such, those results show that GNF scales particularly well with the number of training instances: they tend to not overfit when trained on small databases (e.g. 300W) due to their randomized decision tree nature. When trained on larger corpses (e.g. 300W-LP), their high number of parameters allows to efficiently reduce the training error. Figure \ref{largeposes} shows examples of successful alignment on large poses.

\begin{table*}[ht]
\centering
\begin{minipage}{0.55\textwidth}
\centering
\caption{Comparison with other cascaded regression approaches 300W}
\label{compalignment}
\scalebox{1.0}{
\begin{tabular}{ l | c | c | c | c | c | c }
\hline
\cellcolor{Blue1} & \multicolumn{2}{ c |}{\cellcolor{Blue1}LFPW} & \multicolumn{2}{ c|}{\cellcolor{Blue1}HELEN} & \multicolumn{2}{ c|}{\cellcolor{Blue1}IBUG}\\
\hline
method & $51$ & $68$ & $51$ & $68$ & $51$ & $68$\\
\hline
SDM \cite{xiong2013supervised} &	4.47 &	5.67 & 4.25 &	5.50 & - & 15.4\\
\hline
RCPR \cite{burgos2013robust} & 5.48	& 6.56	& 4.64	& 5.93	& -	& 17.3\\
\hline
DRMF \cite{asthana2013robust}	& 4.40 & 5.80 & 4.60 & 5.80 &	- &	19.8\\
\hline
IFA	\cite{asthana2014incremental} & 6.12 & - & 5.86	& - &	- &	-\\
\hline
CFAN \cite{zhang2014coarse}	& - & 5.44 & - & 5.53 &	- &	-\\
\hline
PO-CR	\cite{tzimiropoulos2015project} & 4.08 & - &	3.90 & - & - & - \\
\hline
L21	\cite{martinez20162} & 3.80 & - & 4.1 & - & 16.3 & -\\
\hline
CSP-dGNF	& \textbf{3.74} & \textbf{4.72} & \textbf{3.59} & \textbf{4.79} & \textbf{10.3} & \textbf{12.0}\\
\hline
\end{tabular}}
\end{minipage}\hfill
\begin{minipage}{0.45\textwidth}
\centering
\caption{Comparison with approaches on AFLW2000-3D}
\label{compalignment3D}
\scalebox{0.8}{
\begin{tabular}{ l | c | c | c | c | c }
\hline
\cellcolor{Blue1} method & \cellcolor{Blue1}$[0,30]$ & \cellcolor{Blue1}$[30,60]$ & \cellcolor{Blue1}$[60,90]$ & \cellcolor{Blue1}avg & \cellcolor{Blue1}std \\
\hline
\multicolumn{6}{ c }{\cellcolor{Blue1}training on 300W}\\
\hline
\scriptsize{RCPR} \cite{burgos2013robust}&4.16&9.88&22.58&12.21&9.43\\
\scriptsize{ESR} \cite{cao2014face}&4.38&10.47&20.31&11.72&8.04\\
\scriptsize{SDM} \cite{xiong2013supervised}&3.56&7.08&17.48&9.37&7.23\\
\scriptsize{CSP-dGNF} &\textbf{2.88}&\textbf{6.33}&\textbf{12.50}&\textbf{7.23}&\textbf{4.87}\\
\hline
\multicolumn{6}{ c }{\cellcolor{Blue1}training on 300W-LP}\\
\hline
\scriptsize{RCPR} \cite{burgos2013robust}&4.26&5.96&13.18&7.80&4.74\\
\scriptsize{ESR} \cite{cao2014face}&4.60&6.70&12.67&7.99&4.19\\
\scriptsize{SDM} \cite{xiong2013supervised}&3.67&4.94&9.76&6.12&3.21\\
\scriptsize{3DDFA} \cite{zhu2015face} &3.78&4.54&7.93&5.42&2.21\\
\scriptsize{3DDFA+SDM} \cite{zhu2015face} &3.43&4.24&7.17&4.94&\textbf{1.97}\\
\scriptsize{CSP-dGNF} &\textbf{2.67}&\textbf{4.19}&\textbf{7.00}&\textbf{4.62}&2.19\\
\hline
\end{tabular}}
\end{minipage}\hfill
\end{table*}

\subsection{Runtime evaluation}\label{realtime}

Table \ref{comptime} shows a runtime evaluation using the settings detailled in Section \ref{hyperparam}. Applying $\mathcal{L}_1$-regularization with truncated gradient on the NN's weights allow to decrease the NN runtime by more than $90\%$. Moreover, using GNF instead of NF allows to reduce the alignment runtime by a factor $100$. Using Sparse NN+GNF, the total runtime is reduced to $12.6$ ms, which allows real-time processing at approximately 80 fps, which is more than most state-of-the-art approaches. This was benchmarked using a loosely-optimized C++ implementation on an $I7-4770$ CPU. 

\begin{table}[h!]
\centering
\caption{Runtime evaluation}
\begin{tabular}{ l | c }
\cellcolor{Blue1} processing step & \cellcolor{Blue1} runtime (ms) \\
\hline
feature extraction & 0.70\\
\hline
NN & 11.7\\
\hline
Sparse NN& 1.51\\
\hline
Regression (NF) & 234.0\\
\hline
Regression (GNF) & 1.42\\
\hline
\hline
Total (NN+NF) & 981.0\\
\hline
Total (Sparse NN+GNF) & \textbf{12.6}\\
\hline
\end{tabular}
\label{comptime}
\end{table}

\section{Conclusion}

In this paper, we introduced a new face alignment framework, that consists in the conjunction of two novel ideas. First, we design a semi-parametric cascade, in which the shape is aligned in the space of a parametric model to provide a precise first guess of the shape. Latter in the cascade, fine-grained deformations are captured with explicit layers. In order to learn each (parametric or explicit) update, we introduced GNF, which contains several improvements over NF: namely a simplified training procedure involving constant prediction nodes (with theoretical guarantees that the trees are covering the regression ranges adequately) as well as a faster greedy evaluation. GNF appears as an ideal predictor for face alignment, as it combines expressivity, fast evaluation, and differentiability that allows a single pass, top-down learning of a NN for dimensionality reduction.

As is, the proposed semi-parametric cascade with sparse NN and GNF allows fast and accurate face alignment for both small, medium and large head poses using baseline features. Future work will consist in incorporating CNN layers for learning low-level representations for face alignment using GNF, as its differentiable nature allows to learn upstream layers in a single pass. For that matter, using data labelled with auxiliary tasks (age, expression or gender prediction) will be considered for pre-training the CNNs.


{\small
\bibliographystyle{ieee}

\begin{thebibliography}{10}\itemsep=-1pt

\bibitem{asthana2013robust}
A.~Asthana, S.~Zafeiriou, S.~Cheng, and M.~Pantic.
\newblock Robust discriminative response map fitting with constrained local
  models.
\newblock In {\em International Conference on Computer Vision and Pattern
  Recognition}, pages 3444--3451, 2013.

\bibitem{asthana2014incremental}
A.~Asthana, S.~Zafeiriou, S.~Cheng, and M.~Pantic.
\newblock Incremental face alignment in the wild.
\newblock In {\em International Conference on Computer Vision and Pattern
  Recognition}, pages 1859--1866, 2014.

\bibitem{belhumeur2013localizing}
P.~N. Belhumeur, D.~W. Jacobs, D.~J. Kriegman, and N.~Kumar.
\newblock Localizing parts of faces using a consensus of exemplars.
\newblock {\em IEEE transactions on pattern analysis and machine intelligence},
  35(12):2930--2940, 2013.

\bibitem{burgos2013robust}
X.~P. Burgos-Artizzu, P.~Perona, and P.~Doll{\'a}r.
\newblock Robust face landmark estimation under occlusion.
\newblock In {\em International Conference on Computer Vision}, pages
  1513--1520, 2013.

\bibitem{cao2014face}
X.~Cao, Y.~Wei, F.~Wen, and J.~Sun.
\newblock Face alignment by explicit shape regression.
\newblock {\em International Journal of Computer Vision}, 107(2):177--190,
  2014.

\bibitem{cootes1998active}
T.~F. Cootes, G.~J. Edwards, and C.~J. Taylor.
\newblock Active appearance models.
\newblock In {\em European Conference on Computer Vision}, pages 484--498,
  1998.

\bibitem{cootes1995active}
T.~F. Cootes, C.~J. Taylor, D.~H. Cooper, and J.~Graham.
\newblock Active shape models-their training and application.
\newblock {\em Computer vision and Image Understanding}, 61(1):38--59, 1995.

\bibitem{dapogny2015pairwise}
A.~Dapogny, K.~Bailly, and S.~Dubuisson.
\newblock Pairwise conditional random forests for facial expression
  recognition.
\newblock In {\em Proceedings of the IEEE International Conference on Computer
  Vision}, pages 3783--3791, 2015.

\bibitem{dollar2009integral}
P.~Doll{\'a}r, Z.~Tu, P.~Perona, and S.~Belongie.
\newblock Integral channel features.
\newblock In {\em British Machine Vision Conference}, 2009.

\bibitem{kontschieder2015deep}
P.~Kontschieder, M.~Fiterau, A.~Criminisi, and S.~Rota~Bulo.
\newblock Deep neural decision forests.
\newblock In {\em International Conference on Computer Vision}, pages
  1467--1475, 2015.

\bibitem{kostinger2011annotated}
M.~K{\"o}stinger, P.~Wohlhart, P.~M. Roth, and H.~Bischof.
\newblock Annotated facial landmarks in the wild: A large-scale, real-world
  database for facial landmark localization.
\newblock In {\em Computer Vision Workshops (ICCV Workshops), 2011 IEEE
  International Conference on}, pages 2144--2151. IEEE, 2011.

\bibitem{langford2009sparse}
J.~Langford, L.~Li, and T.~Zhang.
\newblock Sparse online learning via truncated gradient.
\newblock {\em Journal of Machine Learning Research}, 10(Mar):777--801, 2009.

\bibitem{le2012interactive}
V.~Le, J.~Brandt, Z.~Lin, L.~Bourdev, and T.~S. Huang.
\newblock Interactive facial feature localization.
\newblock In {\em European Conference on Computer Vision}, pages 679--692.
  Springer, 2012.

\bibitem{martinez20162}
B.~Martinez and M.~F. Valstar.
\newblock L 2, 1-based regression and prediction accumulation across views for
  robust facial landmark detection.
\newblock {\em Image and Vision Computing}, 47:36--44, 2016.

\bibitem{ren2014face}
S.~Ren, X.~Cao, Y.~Wei, and J.~Sun.
\newblock Face alignment at 3000 fps via regressing local binary features.
\newblock In {\em International Conference on Computer Vision and Pattern
  Recognition}, pages 1685--1692, 2014.

\bibitem{sagonas2013300}
C.~Sagonas, G.~Tzimiropoulos, S.~Zafeiriou, and M.~Pantic.
\newblock 300 faces in-the-wild challenge: The first facial landmark
  localization challenge.
\newblock In {\em Proceedings of the IEEE International Conference on Computer
  Vision Workshops}, pages 397--403, 2013.

\bibitem{sun2013deep}
Y.~Sun, X.~Wang, and X.~Tang.
\newblock Deep convolutional network cascade for facial point detection.
\newblock In {\em Proceedings of the IEEE Conference on Computer Vision and
  Pattern Recognition}, pages 3476--3483, 2013.

\bibitem{thies2016face2face}
J.~Thies, M.~Zollh{\"o}fer, M.~Stamminger, C.~Theobalt, and M.~Nie{\ss}ner.
\newblock Face2face: Real-time face capture and reenactment of rgb videos.
\newblock {\em Proc. Computer Vision and Pattern Recognition (CVPR), IEEE}, 1,
  2016.

\bibitem{tzimiropoulos2015project}
G.~Tzimiropoulos.
\newblock Project-out cascaded regression with an application to face
  alignment.
\newblock In {\em International Conference on Computer Vision and Pattern
  Recognition}, pages 3659--3667, 2015.

\bibitem{tzimiropoulos2014gauss}
G.~Tzimiropoulos and M.~Pantic.
\newblock Gauss-newton deformable part models for face alignment in-the-wild.
\newblock In {\em International Conference on Computer Vision and Pattern
  Recognition}, pages 1851--1858, 2014.

\bibitem{xiong2013supervised}
X.~Xiong and F.~De~la Torre.
\newblock Supervised descent method and its applications to face alignment.
\newblock In {\em International Conference on Computer Vision and Pattern
  Recognition}, pages 532--539, 2013.

\bibitem{zhang2014coarse}
J.~Zhang, S.~Shan, M.~Kan, and X.~Chen.
\newblock Coarse-to-fine auto-encoder networks for real-time face alignment.
\newblock In {\em European Conference on Computer Vision}, pages 1--16, 2014.

\bibitem{zhang2016learning}
Z.~Zhang, P.~Luo, C.~C. Loy, and X.~Tang.
\newblock Learning deep representation for face alignment with auxiliary
  attributes.
\newblock {\em IEEE transactions on pattern analysis and machine intelligence},
  38(5):918--930, 2016.

\bibitem{zhu2015face}
S.~Zhu, C.~Li, C.~Change~Loy, and X.~Tang.
\newblock Face alignment by coarse-to-fine shape searching.
\newblock In {\em Proceedings of the IEEE Conference on Computer Vision and
  Pattern Recognition}, pages 4998--5006, 2015.

\bibitem{zhu2012face}
X.~Zhu and D.~Ramanan.
\newblock Face detection, pose estimation, and landmark localization in the
  wild.
\newblock In {\em Computer Vision and Pattern Recognition (CVPR), 2012 IEEE
  Conference on}, pages 2879--2886. IEEE, 2012.

\end{thebibliography}

}

\newpage

\section*{Appendix: a lower bound for the depth of randomly initialized NF with constant prediction nodes sampled from a Gaussian distribution for regression}  

In the case of regression, we aim at showing that, provided the tree is deep enough, for each value in the range that shall be covered by the tree, one can find at least one leaf prediction that is close to that value. This way, provided the node initialization is correct and the learning rate hyperparameter is set carefully in order to avoid saturation of the neurons, the error for each training example can theoretically be decreased to less than $\epsilon$. Formally, we aim at proving the following proposition:

\paragraph{proposition:}

If we consider a prediction tree with constant leaf predictions initialized from a gaussian distribution, for each value $y \in [\bar{\delta}_k - \sigma_k, \bar{\delta}_k + \sigma_k]$ there is a probability superior to $1 - \epsilon'$ that there exists at least one leaf prediction $y_l$ such that $|y-y_l|<\epsilon$ if $\mathcal{D}>\mathcal{D}_0$, with

\begin{equation}\label{lowerbound}
\mathcal{D}_0 = \frac{1}{\ln(2)}\ln({\frac{\ln(1-(1-\epsilon')^{\frac{1}{2\sigma_k}})}{\ln(1-\frac{2\epsilon}{\sqrt{2\pi}\sigma_k}e^{-\frac{(\sigma_k+\epsilon)^2}{2\sigma_k^2}})}})
\end{equation}

\paragraph{proof:}

Let $\mathcal{A}$ denote the following event: ``For every value $y \in [\bar{\delta_k}-\sigma_k,\bar{\delta_k}+\sigma_k]$ tree $t$ contains at least one leaf such that the prediction $y_l$ for that leaf satisfies $|y_l-y|<\epsilon$''.

We also define $\mathcal{A}_y$ the event ``For value $y$ there is at least one leaf $l$ of tree $t$, such that $|y_l-y|<\epsilon$''. $p(\mathcal{A})$ can be written as the product integral of probabilities $p(\mathcal{A}_y)$ on interval $[\bar{\delta_k}-\sigma_k,\bar{\delta_k}+\sigma_k]$:

\begin{equation}
 p(\mathcal{A})=\prod\limits_{\bar{\delta_k}-\sigma_k}^{\bar{\delta_k}+\sigma_k}{p(\mathcal{A}_y)^{dy}}
\end{equation}

Which is equivalent to

\begin{equation}\label{appaexpint}
p(\mathcal{A})=\exp(\int_{\bar{\delta_k}-\sigma_k}^{\bar{\delta_k}+\sigma_k}{\ln(p(\mathcal{A}_y)) dy})
\end{equation}

Let's then denote $\bar{\mathcal{A}_y}$ the event: ``for every leaf of tree $t$, $|y_l-y|>\epsilon$. Clearly we have 

\begin{equation}\label{appacontra}
p(\mathcal{A}_y) = 1-p(\bar{\mathcal{A}_y})
\end{equation}

Moreover, as a tree of depth $\mathcal{D}$ shall contain $2^\mathcal{D}$ prediction nodes, we have:

\begin{equation}\label{appaproba}
p(\bar{\mathcal{A}_y}) = p(|y_l-y|>\epsilon)^{2^\mathcal{D}}
\end{equation}

Furthermore, as the leaf predictions $y_l$ are randomly initialized from a gaussian distribution, for one specific leaf node $l$ we can write:

\begin{equation}
p(|y_l-y|>\epsilon) = 1 - \int_{y-\epsilon}^{y+\epsilon}{\frac{1}{\sqrt{2\pi}\sigma_k} e^{-\frac{(z-\delta_k)^2}{2\sigma_k^2}}dz}
\end{equation}

We can use a lower bound of the gaussian function on the interval $[\delta_k-\sigma_k-\epsilon,\delta_k+\sigma_k+\epsilon]$ to provide an upper bound on this probability:

\begin{equation}
p(|y_l-y|>\epsilon) < 1 - \int_{y-\epsilon}^{y+\epsilon}{\frac{1}{\sqrt{2\pi}\sigma_k} e^{-\frac{(\sigma_k+\epsilon)^2}{2\sigma_k^2}}dz}
\end{equation}

thus

\begin{equation}\label{appalowerbound}
p(|y_l-y|>\epsilon) < 1 - \frac{2\epsilon}{\sqrt{2\pi}\sigma_k}e^{-\frac{(\sigma_k+\epsilon)^2}{2\sigma_k^2}}
\end{equation}

Moreover, using Equations \ref{appaexpint}, \ref{appacontra} and \ref{appaproba} we have:

\begin{equation}\label{appa3eq}
p(\mathcal{A})=\exp(\int_{\bar{\delta_k}-\sigma_k}^{\bar{\delta_k}+\sigma_k}{\ln(1-p(|y_l-y|>\epsilon)^{2^\mathcal{D}}) dy})
\end{equation}

Thus, as both $\ln$, $\exp$ and $\int$ are increasing functions, using Equations \ref{appa3eq} and \ref{appalowerbound} provides a lower bound of $p(\mathcal{A})$:

\begin{equation}
p(\mathcal{A})>\exp(\int_{\bar{\delta_k}-\sigma_k}^{\bar{\delta_k}+\sigma_k}{\ln(1-(1 - \frac{2\epsilon}{\sqrt{2\pi}\sigma_k}e^{-\frac{(\sigma_k+\epsilon)^2}{2\sigma_k^2}})^{2^\mathcal{D}}) dy})
\end{equation}

Which we can write

\begin{equation}
p(\mathcal{A})>(1-(1 - \frac{2\epsilon}{\sqrt{2\pi}\sigma_k}e^{-\frac{(\sigma_k+\epsilon)^2}{2\sigma_k^2}})^{2^\mathcal{D}})^{2\sigma_k}
\end{equation}

Thus, a sufficient condition to ensure $p(\mathcal{A})>1-\epsilon'$ (with $\epsilon'$ close to $0$) is to have $\mathcal{D}>\mathcal{D}_0$ with

\begin{equation}
\mathcal{D}_0 = \frac{1}{\ln(2)}\ln({\frac{\ln(1-(1-\epsilon')^{\frac{1}{2\sigma_k}})}{\ln(1-\frac{2\epsilon}{\sqrt{2\pi}\sigma_k}e^{-\frac{(\sigma_k+\epsilon)^2}{2\sigma_k^2}})}})
\end{equation}

Furthermore, when $\epsilon \to 0$, the lower bound depth $\mathcal{D}_0$ is equivalent to:

\begin{equation}
\mathcal{D}_0 {\sim \atop{\mathcal{\epsilon} \to 0}} -\frac{\ln(\epsilon)}{\ln(2)}
\end{equation}

Thus, given the regression range $\sigma_k$ the lower bound depth $\mathcal{D}_0$ grows as the logarithm of the desired ``resolution'' (\textit{i.e.} the inverse of $\epsilon$).

\end{document}